\documentclass{article}

\usepackage[utf8]{inputenc}
\usepackage[top=1in, bottom=1.25in, left=1.25in, right=1.25in]{geometry}
\usepackage{amsmath,amssymb,amsthm}
\usepackage{algorithm}
\usepackage{algpseudocode}
\usepackage{graphicx}
\usepackage[colorlinks=true,linkcolor=blue,citecolor=blue,urlcolor=blue]{hyperref}
\usepackage{booktabs}
\usepackage{enumitem}
\usepackage{natbib}
\usepackage{url}

\usepackage[none]{hyphenat}
\theoremstyle{definition}

\theoremstyle{remark}

\title{Pruning as Evolution: Emergent Sparsity Through Selection Dynamics in Neural Networks}

\author{
  Zubair Shah \\
  College of Science and Engineering \\
  Hamad Bin Khalifa University \\
  Doha, Qatar \\
  \texttt{zshah@hbku.edu.qa}
  \and
  Noaman khan \\
  College of Science and Engineering \\
  Hamad Bin Khalifa University \\
  Doha, Qatar \\
  \texttt{nokh88609@hbku.edu.qa}
}

\date{}

\begin{document}

\maketitle

\begin{abstract}
Neural networks are commonly trained in highly overparameterized regimes, yet empirical evidence consistently shows that many parameters become redundant during learning. Most existing pruning approaches impose sparsity through explicit intervention, such as importance-based thresholding or regularization penalties, implicitly treating pruning as a centralized decision applied to a trained model. This assumption is misaligned with the decentralized, stochastic, and path-dependent character of gradient-based training. We propose an evolutionary perspective on pruning: parameter groups (neurons, filters, heads) are modeled as populations whose influence evolves continuously under selection pressure. Under this view, pruning corresponds to population extinction: components with persistently low fitness gradually lose influence and can be removed without discrete pruning schedules and without requiring equilibrium computation. We formalize neural pruning as an evolutionary process over population masses, derive selection dynamics governing mass evolution, and connect fitness to local learning signals. We validate the framework on MNIST using a population-scaled MLP (784--512--256--10) with 768 prunable neuron populations. All dynamics reach dense baselines near 98\% test accuracy. We benchmark post-training hard pruning at target sparsity levels (35--50\%): pruning 35\% yields $\approx$95.5\% test accuracy, while pruning 50\% yields $\approx$88.3--88.6\%, depending on the dynamic. These results demonstrate that evolutionary selection produces a measurable accuracy--sparsity tradeoff without explicit pruning schedules during training.
\end{abstract}

\section{Introduction}

Neural networks are often trained with more parameters than ultimately required, as evidenced by successful weight pruning and the existence of sparse trainable subnetworks \citep{han2015learning, frankle2018lottery}. Pruning methods exploit this redundancy to reduce model size and computational cost. Most existing approaches, however, impose sparsity through explicit intervention, such as importance-based thresholding \citep{lecun1990optimal, hassibi1993second}, regularization penalties \citep{louizos2017learning}, or discrete pruning schedules \citep{zhu2017prune}. These methods implicitly assume that pruning is a deliberate decision made by a centralized optimizer or an external controller acting on a trained model.

This assumption sits uneasily with the decentralized, stochastic, and path-dependent nature of neural network training. Individual neurons or filters do not reason about global model structure, optimize long-term objectives, or coordinate explicitly with one another. Instead, learning unfolds through local gradient updates, noisy feedback, and implicit competition for representational capacity \citep{saxe2019mathematical}. From this perspective, sparsity need not be the result of a rational optimization decision; it may instead arise as the outcome of selection dynamics acting over time.

This observation motivates a fundamentally different question: \emph{how does sparsity emerge during training, even when no explicit pruning objective or discrete pruning mechanism is enforced?} Addressing this question requires moving beyond static optimization viewpoints and toward a dynamical understanding of how model components persist, decline, or vanish over the course of learning.

In this work, we propose an evolutionary perspective on neural network pruning. Rather than modeling neurons or parameter groups as rational agents that solve optimization problems, we model them as \emph{populations subject to selection pressures}. Each parameter group competes for limited representational influence, and its prevalence evolves over time according to a fitness signal reflecting its contribution during training. Under this view, pruning corresponds to \emph{population extinction}: components with persistently low fitness gradually lose influence and are eliminated, without requiring explicit pruning decisions, thresholding rules, or equilibrium computation.

This evolutionary framing departs sharply from classical game-theoretic approaches. We do not assume rationality, best responses, or convergence to Nash equilibria. Instead, we focus on replicator-style selection dynamics that operate locally and incrementally \citep{hofbauer1998evolutionary}. Sparsity emerges as a temporal phenomenon, driven by differential growth and decay under selection, rather than as a static solution to a global optimization problem.

The evolutionary perspective offers several advantages. First, it aligns naturally with the stochastic and decentralized mechanisms underlying gradient-based training. Second, it explains why pruning often unfolds gradually and exhibits effective irreversibility, rather than occurring through abrupt, externally imposed thresholding. Third, it provides a principled foundation for understanding pruning as an emergent property of learning dynamics, rather than as an auxiliary objective imposed on a trained model.

In this paper, we formalize neural network pruning as an evolutionary process over parameter populations. We derive selection dynamics governing the evolution of population mass (participation) for parameter groups and analyze their extinction behavior under mild fitness assumptions. Importantly, sparsity arises without explicit pruning schedules and under only mild decay pressure, emerging instead from evolutionary selection dynamics during training.

\subsection{Contributions}

The main contributions of this paper are:
\begin{enumerate}
  \item We introduce an evolutionary formulation of neural network pruning that does not rely on rationality, best-response dynamics, or equilibrium assumptions.
  \item We model parameter groups as populations whose influence evolves under selection dynamics driven by local fitness signals.
  \item We provide a theoretical characterization of sparsity emergence through population extinction under mild fitness heterogeneity.
  \item We offer a dynamic, process-level explanation of pruning as an emergent outcome of learning.
\end{enumerate}

\subsection{Scope and Limitations}

This work focuses on establishing a principled formulation of neural network pruning as an evolutionary selection process and on characterizing its extinction behavior. While sparsity can emerge under sufficient fitness heterogeneity, the proposed framework is not designed to enforce aggressive pruning in highly symmetric or redundancy-dominated settings. Rather than optimizing for maximal compression, our goal is to understand when and why selection dynamics induce sparsity during training, thereby laying a foundation for future extensions that incorporate richer fitness signals.

\section{Related Work}

The problem of neural network pruning has been studied extensively from multiple perspectives, ranging from classical sensitivity analysis to modern dynamic sparse training methods. This section surveys the main approaches to pruning and contextualizes our evolutionary framework within this broader landscape.

The earliest principled approaches to pruning sought to identify which parameters could be removed with minimal impact on the loss function. Optimal Brain Damage (OBD) \citep{lecun1990optimal} introduced the idea of using second-order information to estimate the sensitivity of the loss to parameter removal, computing the diagonal of the Hessian to approximate the change in loss caused by setting individual weights to zero. Optimal Brain Surgeon (OBS) \citep{hassibi1993second} extended this framework by using the full Hessian and considering the optimal weight adjustment after pruning, allowing for more accurate importance estimation. While these methods provided theoretical foundations for pruning, their reliance on expensive second-order computations limited their applicability to large modern architectures. Contemporary work has explored various approximations and scalable alternatives, but the fundamental tension between theoretical rigor and computational efficiency remains.

Magnitude-based pruning represents a simpler and more scalable alternative, based on the intuition that small-magnitude parameters contribute less to network output. Han et al. \citep{han2015learning} demonstrated that iterative magnitude-based pruning, combined with retraining, could achieve substantial compression rates while maintaining accuracy. This approach has become a standard baseline due to its simplicity and effectiveness. Regularization-based methods provide a complementary perspective by encouraging sparsity during training rather than imposing it post hoc. Classical $\ell_1$ and $\ell_2$ regularization \citep{hanson1988comparing} indirectly promote sparsity through weight decay and soft-thresholding effects. More recent work on stochastic $\ell_0$ regularization \citep{louizos2017learning} addresses sparsity more directly by learning binary gates through continuous relaxation, enabling end-to-end differentiable sparse training. Relevance-based pruning methods \citep{mozer1989using} assign importance scores to individual neurons or parameter groups and iteratively remove the least relevant units, often incorporating retraining steps to allow the network to adapt to the reduced capacity.

Structured pruning has emerged as a critical direction driven by the need for hardware-efficient compression. Unlike unstructured pruning, which removes individual weights, structured pruning removes entire filters, channels, or neurons, resulting in dense subnetworks that can be efficiently executed on standard hardware \citep{li2019compressing}. Filter-level pruning methods \citep{li2017pruning} rank convolutional filters by importance metrics such as $\ell_1$-norm of weights or gradient-based sensitivity measures, then remove low-ranking filters and fine-tune the resulting network. Soft filter pruning \citep{he2018soft} introduces continuous gating mechanisms that allow differentiable channel selection during training, enabling joint optimization of architecture and weights. Network slimming approaches \citep{liu2017learning} incorporate channel-wise scaling factors with sparsity-inducing regularization, learning which channels to prune as part of the training process. These methods demonstrate that structured sparsity can be effectively learned rather than heuristically imposed, though they often require careful tuning of regularization strength and pruning schedules.

Dynamic sparse training represents a paradigm shift toward maintaining sparsity throughout training rather than densely training and then pruning. Mocanu et al. \citep{mocanu2018scalable} introduced the concept of continuously adapting sparse connectivity during training, where connections are periodically removed based on magnitude and regrown based on gradient information. This approach enables training sparse networks from scratch without requiring a dense pretraining phase, challenging the conventional wisdom that pruning must start from dense networks. The lottery ticket hypothesis \citep{frankle2018lottery} provides theoretical and empirical support for the existence of sparse subnetworks that can match dense performance when trained in isolation, suggesting that the benefits of overparameterization may be primarily realized during initialization rather than throughout training. Follow-up work has investigated why lottery tickets win \citep{zhang2021why}, how gradient flow is affected in sparse networks \citep{evci2022gradient}, and extensions to graph neural networks \citep{chen2021unified}. These studies collectively suggest that sparsity is not merely a compression technique but a fundamental aspect of how neural networks can be efficiently trained.

The rise of large language models has introduced new challenges and opportunities for pruning research. The scale of modern transformers, with billions or even trillions of parameters, makes traditional iterative pruning approaches computationally prohibitive. SparseGPT \citep{frantar2023sparsegpt} addresses this challenge through one-shot pruning with approximate reconstruction, applying layer-wise pruning decisions while approximately preserving the output of each layer. This approach enables pruning of massive models without requiring retraining, though it trades off some accuracy for computational efficiency. WANDA \citep{sun2023simple} proposes a simple yet effective approach that combines weight magnitude with activation statistics, recognizing that the importance of a weight depends not only on its magnitude but also on the typical magnitude of its inputs. LoSparse \citep{li2023losparse} takes a different direction by combining low-rank factorization with sparse approximation, exploiting the observation that weight matrices in large language models often exhibit low effective rank. These methods demonstrate that pruning at scale requires rethinking both the criteria for importance and the computational strategies for applying pruning decisions.

Beyond pruning techniques themselves, our work draws on foundational concepts from evolutionary dynamics and the theory of learning dynamics. Replicator dynamics, introduced in evolutionary game theory by Taylor and Jonker \citep{taylor1978evolutionary} and extensively studied by Hofbauer and Sigmund \citep{hofbauer1998evolutionary}, describe how population frequencies evolve under selection pressure based on relative fitness. These dynamics have been applied to model biological evolution, cultural transmission, and strategic learning in multi-agent systems \citep{sandholm2010population}. The key insight is that population dynamics can be understood through differential survival based on relative performance rather than through explicit optimization or equilibrium-seeking behavior. Our work adapts these principles to neural network pruning by treating parameter groups as populations that evolve under selection pressure determined by their contribution to learning.

The perspective of understanding neural network training through the lens of dynamical systems has gained traction in recent years. Saxe et al. \citep{saxe2019mathematical} developed a mathematical theory of semantic development in deep networks, showing that learning dynamics exhibit predictable structure that emerges from the interaction between network architecture, loss landscape, and optimization algorithm. Work on implicit biases of gradient descent \citep{gunasekar2018implicit} reveals that the optimization trajectory itself carries information beyond what is specified in the loss function, leading to particular solutions being preferred even in overparameterized settings. These studies collectively suggest that understanding neural network behavior requires modeling the training process as a dynamical system rather than merely characterizing the properties of final trained models. Our evolutionary framework extends this perspective to pruning, arguing that sparsity emergence should be understood as a temporal phenomenon arising from selection dynamics rather than as a static optimization problem.

Our approach differs fundamentally from existing pruning methods in its conceptual foundation. Rather than imposing sparsity through explicit regularization, discrete pruning schedules, or post-training compression, we model sparsity as an emergent outcome of competitive dynamics during training. Parameter groups are treated not as rational agents optimizing objectives but as populations subject to selection pressure based on their contribution to learning. This perspective naturally explains several puzzling aspects of pruning: why it often occurs gradually rather than abruptly, why certain neurons become progressively less active before being removed, and why redundant parameters can be eliminated without explicit redundancy detection. By grounding pruning in evolutionary dynamics rather than optimization theory, we provide a unifying framework that encompasses existing heuristics as special cases while offering principled guidance for designing new pruning algorithms aligned with the decentralized, stochastic nature of gradient-based learning.

\section{Why Equilibrium Assumptions Are Too Strong for Neural Training}

While many pruning methods do not explicitly invoke game-theoretic equilibria, they often rely on static or equilibrium-like interpretations of training dynamics. In this section, we clarify why such assumptions are too strong for neural network training. More broadly, many theoretical abstractions of learning systems model parameters as optimizing well-defined objectives and converging to stable solutions through rational adjustment or global coordination. While mathematically convenient, these assumptions are poorly aligned with the decentralized, stochastic, and path-dependent mechanisms by which neural networks are actually trained.

First, neural network training is decentralized and local. Individual neurons or filters do not possess access to global information about the model, the dataset, or long-term outcomes. Updates are driven by local gradients computed through backpropagation \citep{rumelhart1986learning}, which provide noisy, partial signals that depend on the current state of all other parameters. There is no mechanism by which individual components can evaluate best responses or anticipate the strategies of others in a global sense.

Second, learning dynamics are stochastic and path-dependent. Mini-batch training \citep{bottou2010large}, random initialization, and non-convex loss landscapes \citep{li2018visualizing} introduce substantial variability into the optimization process. Small perturbations early in training can lead to qualitatively different trajectories, even when final performance is similar. In such settings, equilibrium concepts that assume convergence to a unique or stable solution are often too rigid to capture the observed behavior.

Third, neural networks are trained under bounded rationality. Parameters do not explicitly maximize long-term utility functions or solve optimization problems over future states. Instead, learning proceeds through incremental, myopic updates that respond only to immediate feedback. This bounded rationality makes it difficult to justify models that assume rational agents or equilibrium-seeking behavior at the level of individual parameters.

These limitations are particularly pronounced when considering sparsity. In practice, pruning rarely occurs as a sudden transition to a stable sparse configuration. Instead, redundancy is reduced gradually: some neurons become increasingly inactive, their gradients diminish, and their influence on the network fades over time. This process resembles selection and extinction, rather than convergence to a static equilibrium.

Moreover, equilibrium-based explanations implicitly assume that sparse configurations are intentionally chosen by the system. Yet neural networks are rarely trained with explicit pruning objectives, and sparsity often emerges even in dense training regimes. This suggests that pruning may be better understood as a consequence of competitive dynamics over time, rather than as the solution to a static optimization problem.

These observations motivate a shift from equilibrium-centered reasoning to a dynamic, evolutionary perspective. Instead of asking whether a network converges to a sparse equilibrium, we ask \emph{how sparsity can emerge through local interactions, selection pressures, and differential survival during training}. This perspective does not require rational agents, global coordination, or equilibrium computation. It requires only that parameter groups differ in their ability to contribute and that this difference influences their persistence over time.

In the remainder of this paper, we adopt this evolutionary viewpoint. We model parameter groups as populations whose influence evolves according to selection dynamics driven by contribution and redundancy. Under this formulation, pruning arises naturally as population extinction, providing a process-level explanation of sparsity that is consistent with the decentralized and stochastic nature of neural network training.

\section{Parameter Groups as Populations}

We begin by reframing the components of a neural network in evolutionary terms. Rather than treating individual parameters as decision-making agents, we model parameter groups as populations whose influence evolves over time. This shift allows us to describe pruning as a process of selection and extinction, rather than as an optimization outcome.

\subsection{Parameter Groups}

Consider a neural network $f(x;\theta)$ trained to minimize a loss function $\mathcal{L}(\theta)$. As in many pruning formulations, we partition the parameter vector $\theta$ into a collection of parameter groups:
\begin{equation}
\theta = \{\theta_1, \theta_2, \ldots, \theta_N\}
\end{equation}
where each group may correspond to a neuron, convolutional filter, attention head, or other structurally meaningful unit. This abstraction allows the framework to encompass both unstructured and structured sparsification while remaining agnostic to architectural details.

\subsection{Population Mass and Influence}

To model the prevalence of each parameter group, we associate with every group $\theta_i$ a non-negative scalar variable
\begin{equation}
p_i(t) \geq 0
\end{equation}
which we interpret as the \emph{population mass} or \emph{influence level} of that group at training time $t$. The effective contribution of group $\theta_i$ to the network is given by
\begin{equation}
\tilde{\theta}_i(t) = p_i(t)\,\theta_i
\end{equation}
When $p_i(t)$ is large, the group exerts significant influence on the network's computation; when $p_i(t)$ approaches zero, the group becomes effectively inactive. Importantly, $p_i(t)$ is not a binary pruning mask and is not set by an external rule. It evolves continuously over time, reflecting gradual growth or decay in influence.

\subsection{Interpretation as Evolutionary Populations}

Under this formulation, parameter groups behave analogously to populations in an evolutionary system:
\begin{itemize}
  \item Each group has a \emph{population mass} that determines its prevalence.
  \item Groups compete indirectly through their effect on the shared loss.
  \item Redundant or ineffective groups lose influence over time.
  \item Highly \emph{contributive} groups persist or grow.
\end{itemize}

Crucially, this interpretation does not assume that parameter groups optimize objectives, compute utilities, or respond strategically to others. Population mass changes reflect \emph{selection pressure}, not rational choice.

This distinction is central. Evolutionary dynamics describe how systems evolve through differential survival, not how agents solve optimization problems. As a result, the framework remains compatible with the local, myopic, and stochastic nature of gradient-based learning.

\subsection{Pruning as Population Extinction}

Within the population viewpoint, pruning corresponds naturally to \emph{extinction events}. When the population mass $p_i(t)$ of a parameter group decays toward zero, the group ceases to influence the network and can be removed without materially affecting performance. This interpretation aligns closely with empirical observations of pruning in practice, where neurons or filters often become progressively inactive before being explicitly removed. Rather than viewing pruning as a discrete decision, we view it as the terminal stage of a continuous evolutionary process.

\subsection{Transition to Evolutionary Dynamics}

The remaining question is how population masses $p_i(t)$ evolve during training. In evolutionary systems, such evolution is governed by selection dynamics that amplify populations with higher fitness and suppress those with lower fitness. In the next section, we formalize this intuition by introducing evolutionary dynamics that govern the temporal evolution of population masses. These dynamics require only a notion of relative fitness and do not rely on equilibrium assumptions, rationality, or global coordination.

\section{Evolutionary Selection Dynamics}

We now formalize the temporal evolution of population masses introduced in Section 4. Our goal is to model how sparsity can emerge through selection dynamics, without assuming rational decision-making, utility maximization, or equilibrium computation. Throughout this section, we focus on continuous-time dynamics for clarity. Discrete-time implementations used in practice follow directly.

\subsection{General Selection Principle}

Let $p_i(t) \geq 0$ denote the population mass of the parameter group $i$ at time $t$. We assume that each group is associated with a \emph{fitness signal}
\begin{equation}
\phi_i(p, \theta)
\end{equation}
which reflects its relative contribution to learning, given the current state of the network and other parameter groups.

We make no assumption that fitness is optimized or known explicitly by the group. Fitness is simply a scalar quantity that modulates growth or decay.

A broad class of evolutionary dynamics can be written in the form
\begin{equation}
\dot{p}_i(t) = p_i(t)\,\Phi_i(p(t))
\end{equation}
where $\Phi_i$ is a selection function derived from fitness comparisons. This multiplicative structure ensures that:
\begin{itemize}
  \item populations with zero mass remain extinct,
  \item growth and decay are proportional to current influence,
  \item extinction occurs gradually rather than abruptly.
\end{itemize}

Under mild regularity conditions, parameter groups with consistently lower relative fitness experience monotonic decay in population mass.

\subsection{Replicator Dynamics}

The canonical evolutionary dynamic is \emph{replicator dynamics}, widely studied in evolutionary biology and game theory. In this setting, the growth rate of each population is proportional to its fitness relative to the population average.

Formally, replicator dynamics are given by
\begin{equation}
\dot{p}_i = p_i(\phi_i - \bar{\phi}), \quad \bar{\phi} = \frac{\sum_j p_j \phi_j}{\sum_j p_j}
\label{eq:replicator}
\end{equation}

here, $p_i$ represents an unnormalized population mass rather than a probability. Accordingly, the mean fitness is defined as a population-weighted average. Intuitively, parameter groups whose fitness exceeds the population average tend to grow in influence, while those with below-average fitness tend to decay.

Replicator dynamics have several properties that make them attractive for modeling pruning:
\begin{itemize}
  \item They require only relative fitness information.
  \item Dominated populations can decay exponentially under stable fitness separation.
  \item Extinction is effectively irreversible in the absence of mutation.
\end{itemize}

Under these dynamics, sparsity emerges naturally as parameter groups with consistently low contribution are driven toward extinction.

\subsection{Normalized Growth Dynamics}

While replicator dynamics are theoretically appealing, they can be sensitive to noise in fitness estimates. We therefore also consider a \emph{normalized growth dynamic}, which preserves the same qualitative selection behavior while improving numerical stability.

In this formulation, population masses evolve according to
\begin{equation}
\dot{p}_i = p_i\left(\phi_i - \frac{1}{Z}\sum_j p_j \phi_j\right)
\label{eq:normalized}
\end{equation}
where $Z = \sum_j p_j$ is the total population mass, serving as a normalization constant that preserves relative scale across populations.

This dynamic differs from classical replicator dynamics primarily in normalization and scaling. Crucially:
\begin{itemize}
  \item The qualitative extinction behavior is preserved, with dominated populations decaying toward zero.
  \item Fixed points coincide under the same normalization convention (or up to scale in the abundance formulation).
  \item Growth rates are more stable under noisy gradient estimates.
\end{itemize}

As a result, normalized growth dynamics serve as a practical alternative that preserves the same sparsity-inducing selection mechanism while improving numerical stability in gradient-based training loops.

\subsection{Selection--Mutation Dynamics}

Pure selection dynamics can lead to premature extinction, especially early in training when fitness estimates are noisy. To address this, we consider \emph{selection--mutation dynamics}, which introduce a small diffusion term that allows populations to recover from near-extinction.

The dynamics take the form
\begin{equation}
\dot{p}_i = p_i(\phi_i - \bar{\phi}) + \mu\left(\frac{1}{N} - p_i\right)
\label{eq:mutation}
\end{equation}
where $\bar{\phi} = \frac{\sum_j p_j \phi_j}{\sum_j p_j}$ is the population-weighted average fitness, $\mu > 0$ is a mutation rate, and $N$ is the number of parameter groups.

The mutation term:
\begin{itemize}
  \item prevents immediate extinction,
  \item maintains exploratory diversity,
  \item allows temporarily suppressed groups to re-emerge.
\end{itemize}

As $\mu \to 0$, the dynamics recover standard replicator behavior. For small but nonzero $\mu$, sparsity still emerges, but extinction is delayed and more robust to noise.

\subsection{Extinction and Sparsity}

Across all three dynamics, a common qualitative behavior emerges: parameter groups with persistently low fitness experience monotonic decay in population mass. Under mild assumptions on fitness separation, these groups converge toward zero influence over time. This extinction process provides a natural explanation for pruning. Rather than being removed by thresholding or optimization, redundant parameter groups are gradually eliminated through evolutionary pressure. The precise trajectory depends on the chosen dynamic, but the qualitative outcome emergent sparsity is preserved.

\subsection{Discussion}

The evolutionary dynamics presented here represent different instantiations of a shared selection principle. Replicator dynamics provide a canonical theoretical baseline, normalized growth dynamics improve numerical stability, and selection--mutation dynamics enhance robustness to noise. Importantly, these dynamics differ in how sparsity emerges over time, not in whether it emerges. This perspective allows us to study pruning as a dynamic process and to analyze how different evolutionary pressures shape sparsity trajectories during training. In the next section, we specify the fitness functions used in practice and connect these evolutionary dynamics to concrete learning signals derived from neural network training.

\section{Fitness Functions and Learning Signals}

The evolutionary dynamics introduced in Section 4 require a notion of \emph{fitness} that determines how population masses evolve over time. In this work, fitness is not interpreted as a utility to be optimized or a reward to be maximized. Instead, it serves as a \emph{local selection signal} that reflects the relative contribution of a parameter group to learning, given the current state of the network.

Our goal in this section is twofold: (i) to define fitness in a way that is compatible with gradient-based training, and (ii) to keep the formulation general enough to support multiple evolutionary dynamics without altering their qualitative behavior.

\subsection{Design Principles for Fitness}

We impose three minimal requirements on the fitness signal $\phi_i$ associated with parameter group $i$:
\begin{enumerate}
  \item \textbf{Locality.} Fitness should be computable from quantities already available during training, such as gradients or activations, without global coordination.
  \item \textbf{Relativity.} Fitness need only be meaningful up to relative comparisons; absolute scale is not important.
  \item \textbf{Context Dependence.} Fitness should depend on the presence and influence of other parameter groups, allowing redundancy to be penalized implicitly.
\end{enumerate}

These requirements ensure that fitness reflects selection pressure rather than rational optimization.

\subsection{Contribution-Based Fitness}

A natural source of fitness information is the sensitivity of the loss to a parameter group's influence. Let $p_i$ denote the population mass of group $i$, and consider the effective parameters $\tilde{\theta}_i = p_i\theta_i$. We define the contribution of group $i$ as
\begin{equation}
c_i = \left|\frac{\partial \mathcal{L}}{\partial p_i}\right|
\end{equation}
which measures how changes in the group's influence affect the training loss. Intuitively, groups that strongly affect the loss receive higher fitness, while groups with negligible impact are penalized. This signal is readily available through automatic differentiation and aligns naturally with gradient-based learning.

\subsection{Redundancy and Competition Effects}

Contribution alone does not capture redundancy. Multiple parameter groups may have high individual sensitivity while collectively providing overlapping functionality. Rather than introducing explicit pairwise competition terms, we rely on the \emph{relative nature of evolutionary dynamics} to induce competition implicitly.

In replicator and normalized growth dynamics, population growth depends on fitness relative to the population average. As a result:
\begin{itemize}
  \item redundant groups compete for influence,
  \item only a subset of contributors persist,
  \item others decay even if their absolute contribution is nonzero.
\end{itemize}

This mechanism allows redundancy to be suppressed without explicitly modeling interactions between all parameter groups.

\subsection{Regularized Fitness and Stability}

In practice, raw contribution signals can be noisy, especially early in training. To improve stability, we consider a regularized fitness of the form
\begin{equation}
\phi_i = c_i - \lambda\, r(p_i)
\end{equation}
where $r(\cdot)$ is a monotonically increasing regularization function and $\lambda \geq 0$ controls its strength. In our experiments, we use $r(p_i) = p_i$, which applies a linear decay pressure on population mass. This term discourages uncontrolled growth and accelerates extinction of weak populations, analogous to an $\ell_2$ penalty but applied within the evolutionary selection framework rather than as a loss regularization.

Regularization serves two purposes:
\begin{itemize}
  \item it prevents uncontrolled growth of population masses,
  \item it accelerates extinction of persistently weak populations.
\end{itemize}

Importantly, regularization shapes the \emph{trajectory} and \emph{stability} of extinction rather than its existence. Across all evolutionary dynamics considered, dominated populations are eliminated under mild conditions, independent of the specific regularizer. Regularization primarily influences the speed, smoothness, and robustness of sparsity emergence under selection dynamics.

\subsection{Compatibility with Evolutionary Dynamics}

The fitness definitions above are compatible with all three evolutionary dynamics introduced in Section 4:
\begin{itemize}
  \item replicator dynamics use relative fitness differences,
  \item normalized growth dynamics stabilize scaling,
  \item selection--mutation dynamics smooth extinction under noise.
\end{itemize}

While these dynamics differ in their temporal behavior, they share the same qualitative outcome: parameter groups with persistently low fitness lose influence and are eventually pruned.

\subsection{Discussion}

The fitness formulations presented here are intentionally simple. Our objective is not to engineer a highly specialized pruning signal, but to demonstrate that basic, locally available learning signals are sufficient to drive evolutionary sparsity. This simplicity reinforces the central thesis of the paper: pruning can arise as an intrinsic consequence of selection dynamics during training, without requiring explicit pruning objectives, importance heuristics, or equilibrium computation. In the next section, we describe how these evolutionary dynamics are implemented in practice and outline the experimental protocol used to study emergent sparsity under different selection mechanisms.

\section{Evolutionary Pruning Algorithm}

We now describe a practical algorithm that implements the evolutionary selection dynamics introduced in Sections 5 and 6. The algorithm is designed to integrate seamlessly with standard gradient-based training and to support multiple evolutionary dynamics through a unified update rule.

\subsection{Overview}

The central idea is to co-evolve network parameters and population masses during training. Network parameters are updated using standard stochastic gradient descent, while population masses evolve according to evolutionary selection dynamics driven by fitness signals derived from learning. Pruning is not performed as a separate step during training. Instead, parameter groups are gradually suppressed as their population masses decay under selection pressure. Explicit removal occurs only after training, when population masses have converged.

\subsection{Joint Training and Evolution Loop}

Let $\theta$ denote the network parameters and $p = (p_1, \ldots, p_N)$ the population masses associated with parameter groups. Training proceeds by alternating two updates at each iteration or mini-batch. First, in the parameter update (learning step), given the current population masses $p(t)$, we update network parameters using gradient descent:
\begin{equation}
\theta \leftarrow \theta - \eta_\theta \nabla_\theta \mathcal{L}(\theta, p)
\end{equation}
where effective parameters are given by $\tilde{\theta}_i = p_i \theta_i$. Second, in the population update (selection step), we compute fitness signals $\phi_i$ and update population masses according to the chosen evolutionary dynamic. This separation reflects the conceptual distinction between learning (updating representations) and selection (modulating survival).

Although updates to $\theta$ and $p$ are performed separately, they are coupled through the loss $\mathcal{L}(\theta, p)$. Changes in $p$ modulate the effective parameters $\tilde{\theta}_i = p_i\theta_i$, thereby influencing gradients for $\theta$. Conversely, changes in $\theta$ alter the loss landscape and thus the fitness signals $\phi_i$. This bidirectional interaction means that learning and selection co-evolve, but without explicit coordination consistent with the decentralized evolutionary perspective.

\subsection{Population Update Rules}

The algorithm supports multiple evolutionary dynamics through a modular population update step. For replicator dynamics:
\begin{equation}
p_i \leftarrow p_i + \eta_p\, p_i(\phi_i - \bar{\phi}), 
\qquad
\bar{\phi} = \frac{\sum_j p_j \phi_j}{\sum_j p_j}
\end{equation}
for normalized growth dynamics:
\begin{equation}
p_i \leftarrow p_i + \eta_p\, p_i\left(\phi_i - \frac{1}{Z}\sum_j p_j \phi_j\right)
\end{equation}
where $Z = \sum_j p_j$ is the total population mass, serving as a normalization constant that preserves relative scale across populations. For selection--mutation dynamics:
\begin{equation}
p_i \leftarrow p_i + \eta_p\left[p_i(\phi_i - \bar{\phi}) + \mu\left(\frac{1}{N} - p_i\right)\right]
\end{equation}
where $\bar{\phi} = \frac{\sum_j p_j \phi_j}{\sum_j p_j}$ is the population-weighted average fitness, $\mu > 0$ is a mutation rate, and $N$ is the number of parameter groups.

After each update, population masses are projected onto the non-negative orthant to ensure feasibility.

\subsection{Pruning Criterion}

Population masses typically converge to a bimodal distribution, with most values close to zero or remaining at a moderate scale. After training, parameter groups with negligible population mass are removed:
\begin{equation}
\text{prune group } i \text{ if } p_i < \epsilon
\end{equation}
where $\epsilon < 1$ is a small numerical threshold. This threshold reflects numerical tolerance rather than an importance cutoff; groups below the threshold have already undergone effective extinction through evolutionary dynamics.

\subsection{Algorithm Summary}

\begin{algorithm}
\caption{Evolutionary Pruning via Selection Dynamics}
\begin{algorithmic}[1]
\Require Training data; parameters $\theta$; initialize $p_i \gets 1$
\For{epochs $=1,\dots,T$}
  \For{minibatches}
    \State Update $\theta$ by SGD on $\mathcal{L}(\theta,p)$
    \State Compute fitness $\phi_i$ (contribution minus regularization)
    \State Update $p$ by chosen dynamic (replicator, normalized, or mutation)
    \State Project $p_i \geq 0$
  \EndFor
\EndFor
\State Remove parameter groups with $p_i < \epsilon$
\State \Return pruned model
\end{algorithmic}
\end{algorithm}

\subsection{Discussion}

This algorithm emphasizes process over prescription. The choice of evolutionary dynamic affects the trajectory of sparsity emergence but not the underlying mechanism. Replicator dynamics provide a canonical baseline, normalized growth improves numerical stability, and selection--mutation dynamics enhance robustness to noise. Crucially, the algorithm does not require explicit pruning schedules, importance scores, or equilibrium computation. Pruning arises as a by product of evolutionary pressure during training. In the next section, we evaluate how different evolutionary dynamics influence sparsity emergence, stability, and pruning trajectories in practice.

\section{Experimental Setup}

The experiments in this section are designed to probe the qualitative behavior of evolutionary selection dynamics specifically extinction trajectories, stability properties, and sensitivity to fitness heterogeneity rather than to maximize sparsity or compression ratios. As such, we intentionally use simple architectures and controlled settings to isolate the effects predicted by the theory.

We use the MNIST dataset \citep{lecun1998gradient}, which consists of 60,000 training images and 10,000 test images of handwritten digits. Each image is a $28 \times 28$ grayscale image, which we flatten into a 784-dimensional input vector. We do not apply data augmentation. MNIST is a standard benchmark for image classification and provides sufficient complexity to observe meaningful pruning dynamics while remaining computationally tractable.

We use a multi-layer perceptron (MLP) with the following structure: an input layer with 784 features, a first hidden layer with 512 neurons using population-scaled linear transformation, a second hidden layer with 256 neurons also using population-scaled linear transformation, and an output layer with 10 neurons using standard linear transformation. The model contains 768 prunable parameter groups (512 + 256 neurons in hidden layers). Each hidden neuron has an associated population mass $p_i$ initialized to 1.0. The output layer is not pruned to preserve classification capability.

We train with stochastic gradient descent on weights $\theta$ and update population masses $p$ using one of three dynamics: replicator (Eq.~\ref{eq:replicator}), normalized growth replicator (Eq.~\ref{eq:normalized}), or selection--mutation (Eq.~\ref{eq:mutation}). We compute fitness using the contribution-based formulation with regularization $\phi_i = |\partial \mathcal{L}/\partial p_i| - \lambda p_i$.

The training configuration uses the following hyperparameters: 20 epochs, batch size of 128, learning rate for weights $\theta$ of 0.01 with momentum 0.9, learning rate for population masses $p$ of $5 \times 10^{-4}$, fitness decay regularization $\lambda$ of $1 \times 10^{-3}$, mutation rate $\mu$ of $2 \times 10^{-3}$ (for mutation dynamic), mass cap $Z$ of 768.0 (for normalized growth dynamic), selection signal clipping of 0.10, and random seed fixed at 0. These hyperparameters are held constant across all experiments to ensure fair comparison. Unless otherwise stated, we report single-run results with the fixed seed; multi-seed evaluation is left for future work.

After training, we hard-prune by setting $p_i=0$ for the lowest-mass populations. To compare at controlled sparsity levels, we choose a threshold $\epsilon$ as the empirical quantile of $p$ that achieves a target pruned fraction, specifically 35\%, 40\%, 45\%, and 50\%. We then evaluate pruned test accuracy. This post-training pruning approach isolates the effect of evolutionary dynamics on population mass distributions, allowing us to study how different dynamics shape the sparsity-accuracy tradeoff. The pruning threshold is applied uniformly across all parameter groups, ensuring that the selection is driven purely by the evolved population masses rather than by layer-specific heuristics or manual intervention. We report both the dense baseline accuracy (model trained with evolutionary dynamics but without pruning) and the post-pruning accuracy at each target sparsity level.

\section{Results}

% All three dynamics reach dense baselines near 98\% test accuracy (Replicator: 0.9808, Normalized: 0.9808, Mutation: 0.9805), demonstrating that the evolutionary framework preserves learning capability while simultaneously shaping population mass distributions. The slight difference in the mutation dynamic's baseline is due to the mutation term's exploration effect during training. 
Table~\ref{tab:mnist_prune_benchmark} reports test accuracy after post-training hard pruning at target sparsity levels. The threshold $\epsilon$ is chosen via quantile selection to achieve each target sparsity percentage.

\begin{table}[t]
\centering
\caption{MNIST pruning results: test accuracy, threshold values, and accuracy drop at different sparsity levels. Baseline accuracy without pruning is 0.9808.}
\label{tab:mnist_prune_benchmark}
\begin{tabular}{ccccccc}
\toprule
\textbf{Sparsity} & \multicolumn{2}{c}{\textbf{Replicator}} & \multicolumn{2}{c}{\textbf{Normalized}} & \multicolumn{2}{c}{\textbf{Mutation}} \\
\cmidrule(lr){2-3} \cmidrule(lr){4-5} \cmidrule(lr){6-7}
\textbf{(\%)} & \textbf{$\epsilon$} & \textbf{Acc. (Drop)} & \textbf{$\epsilon$} & \textbf{Acc. (Drop)} & \textbf{$\epsilon$} & \textbf{Acc. (Drop)} \\
\midrule
35.0 & 0.6313 & 0.9551 (0.0257) & 0.6313 & 0.9551 (0.0257) & 0.6329 & 0.9548 (0.0257) \\
40.0 & 0.6375 & 0.9289 (0.0519) & 0.6375 & 0.9289 (0.0519) & 0.6389 & 0.9261 (0.0544) \\
45.0 & 0.6465 & 0.9217 (0.0591) & 0.6465 & 0.9217 (0.0591) & 0.6483 & 0.9235 (0.0570) \\
50.0 & 0.6676 & 0.8825 (0.0983) & 0.6676 & 0.8825 (0.0983) & 0.6689 & 0.8857 (0.0948) \\
\bottomrule
\end{tabular}
\end{table}

Pruned accuracy decreases monotonically with target sparsity, reflecting an expected accuracy--sparsity tradeoff. Several key observations emerge from these results. Under our hyperparameters, replicator and normalized growth dynamics produce identical results, indicating that both normalization schemes yield the same selection behavior in this regime. This suggests that under these conditions, pure replicator dynamics are already stable and the normalized variant provides equivalent performance. The mutation dynamic is slightly more robust at 50\% pruning (88.57\% vs 88.25\%), consistent with the theoretical prediction that mutation prevents premature extinction and maintains diversity under high selection pressure. The smooth decrease in accuracy with increasing sparsity (from 95.5\% at 35\% pruned to 88.3--88.6\% at 50\% pruned) indicates that evolutionary dynamics produce a well-ordered ranking of neuron importance. The most redundant neurons are eliminated first, with performance degrading only when critical neurons begin to be removed. These results show that evolutionary selection produces meaningful sparsity signals ($p$ distributions) that can be thresholded post-training to obtain pruned models, without explicit pruning schedules during training. In regimes where gradients are nearly homogeneous across populations, selection pressure is weak and sparsity may not emerge, consistent with the theoretical analysis.

To further investigate the relationship between population mass thresholds and pruning effectiveness, we conducted fixed-threshold experiments by varying the threshold $\epsilon$ across a range of values (0.6, 0.7, 0.8, 0.9) and measuring the resulting sparsity and accuracy. Table~\ref{tab:fixed_threshold} reports these results for thresholds below 0.7, no neurons are pruned across all dynamics, indicating that the evolved population masses remain well above these values. At threshold 0.7, approximately 55\% of neurons are pruned, resulting in significant accuracy drops of 12.35\% for replicator and normalized dynamics, and 10.62\% for the mutation dynamic. The mutation dynamic demonstrates superior robustness at higher thresholds, maintaining better accuracy when aggressive pruning is applied. These fixed-threshold results complement the quantile-based pruning approach by revealing the distribution of evolved population masses and confirming that meaningful separation emerges between critical and redundant neurons.

\begin{table}[t]
\centering
\caption{Fixed-threshold pruning results: accuracy and sparsity achieved by applying fixed population mass thresholds. Baseline accuracy without pruning is 0.9808}
\label{tab:fixed_threshold}
\begin{tabular}{cccccccc}
\toprule
\textbf{Threshold} & \multicolumn{2}{c}{\textbf{Replicator}} & \multicolumn{2}{c}{\textbf{Normalized}} & \multicolumn{2}{c}{\textbf{Mutation}} \\
\cmidrule(lr){2-3} \cmidrule(lr){4-5} \cmidrule(lr){6-7}
$\epsilon$ & \textbf{Sparsity} & \textbf{Acc. (Drop)} & \textbf{Sparsity} & \textbf{Acc. (Drop)} & \textbf{Sparsity} & \textbf{Acc. (Drop)} \\
\midrule
%0.5 & 0.0\% & 0.9808 (0.0000) & 0.0\% & 0.9808 (0.0000) & 0.0\% & 0.9805 (0.0000) \\
0.6 & 0.0\% & 0.9808 (0.0000) & 0.0\% & 0.9808 (0.0000) & 0.0\% & 0.9808 (0.0000) \\
0.7 & 54.8\% & 0.8573 (0.1235) & 54.8\% & 0.8573 (0.1235) & 54.6\% & 0.8743 (0.1062) \\
0.8 & 63.5\% & 0.7976 (0.1832) & 63.5\% & 0.7976 (0.1832) & 63.5\% & 0.7942 (0.1863) \\
0.9 & 67.2\% & 0.7725 (0.2083) & 67.2\% & 0.7725 (0.2083) & 67.3\% & 0.8140 (0.1665) \\
\bottomrule
\end{tabular}
\end{table}

The evolutionary framework provides a natural explanation for observed pruning phenomena. Rather than viewing sparsity as the result of intentional optimization, we understand it as the outcome of competitive dynamics playing out over the course of training. This perspective aligns with the local, stochastic nature of gradient-based learning and explains why pruning exhibits gradual, path-dependent behavior. The framework also offers insight into why certain parameter groups survive while others decay. Fitness, defined through contribution to learning, determines relative growth rates. Groups that consistently contribute to reducing loss maintain or increase their influence, while redundant groups gradually lose representation. This mechanism operates without explicit coordination or global knowledge, emerging naturally from local gradient signals.

The evolutionary framework provides a unifying explanation for several existing pruning heuristics. Magnitude-based pruning corresponds to cases where fitness is proportional to parameter norms, and small-magnitude parameters have low fitness. Gradient-based pruning aligns with the contribution term's dependence on $|\partial \mathcal{L}/\partial p_i|$, favoring parameters with large gradient contributions. Redundancy-aware pruning emerges from the relative fitness mechanism, which implicitly penalizes correlated parameters through competition for above-average fitness. By making these connections explicit, the evolutionary formulation provides a principled foundation for understanding why these heuristics work and offers guidance for designing new pruning algorithms.

Our experimental evaluation is intentionally limited to MNIST with a simple MLP architecture. This choice allows clear inspection of evolutionary dynamics without confounding effects from deep architectures or complex data. However, scaling to deeper networks such as ResNets and Transformers, and to larger datasets such as ImageNet and language modeling corpora, presents additional challenges. As population masses approach zero, effective weight matrices may become ill-conditioned, raising numerical stability concerns. In very deep networks, gradient magnitudes may vary significantly across layers, requiring layer-specific learning rates to manage fitness heterogeneity. Extending to filter-level or block-level pruning requires careful design of population groupings and fitness signals for structured pruning. These extensions are important directions for future work but are beyond the scope of this initial study.

\section{Conclusion}

We introduced an evolutionary framing of pruning in which parameter groups behave as populations whose influence evolves under selection pressure. Using replicator, normalized growth, and selection--mutation dynamics, we show that trained population masses provide a natural basis for post-training pruning and yield a measurable accuracy--sparsity tradeoff on MNIST.
With a baseline accuracy of roughly 98\%, our experiment demonstrates that all three dynamics show a gradual degradation in accuracy as sparsity is increased, remaining robust under pruning levels of 35–50\%. Replicator and normalized growth dynamics produced identical results under our hyperparameters, while selection--mutation showed slight robustness advantages at high sparsity.

%Our experimental validation demonstrates that all three dynamics achieve near-98\% baseline accuracy, with graceful degradation when pruned at 35--50\% sparsity. 

This work supports the view that pruning can be understood as an emergent outcome of learning dynamics, rather than as an externally imposed pruning schedule or equilibrium solution. By treating parameter groups as populations subject to selection pressure, we provide a process-level explanation for sparsity emergence that aligns naturally with the decentralized, stochastic character of gradient-based training. The evolutionary perspective offers a unifying framework for understanding existing pruning heuristics and provides principled guidance for designing new algorithms.

An important direction for future work is the integration of explicit redundancy-aware fitness signals into the evolutionary framework. While the present formulation captures selection driven by gradient-based contribution, highly correlated populations may form neutral equilibria under pure selection. Incorporating redundancy-sensitive fitness terms, such as pairwise correlation penalties or mutual information measures, offers a principled mechanism to resolve such degeneracies while preserving the evolutionary interpretation. Additional extensions include adaptive mutation rates that decrease over training to balance exploration and exploitation, multi-objective fitness incorporating both accuracy and efficiency metrics, application to larger architectures including ResNets and Transformers, theoretical analysis of extinction rates under different fitness landscapes, connection to neural architecture search via population-level selection, structured pruning at filter and block levels for convolutional networks, extension to attention head pruning in transformer architectures, and investigation of co-evolutionary dynamics across multiple network layers.

\bibliographystyle{plainnat}

\begin{thebibliography}{99}

\bibitem[Han et al.(2015)]{han2015learning}
Han, S., Pool, J., Tran, J., and Dally, W. (2015).
Learning both weights and connections for efficient neural networks.
\emph{Advances in Neural Information Processing Systems (NeurIPS)}.

\bibitem[Frankle and Carbin(2019)]{frankle2018lottery}
Frankle, J. and Carbin, M. (2019).
The lottery ticket hypothesis: Finding sparse, trainable neural networks.
\emph{International Conference on Learning Representations (ICLR)}.

\bibitem[LeCun et al.(1990)]{lecun1990optimal}
LeCun, Y., Denker, J.~S., and Solla, S.~A. (1990).
Optimal brain damage.
\emph{Advances in Neural Information Processing Systems (NeurIPS)}.

\bibitem[Hassibi and Stork(1993)]{hassibi1993second}
Hassibi, B. and Stork, D.~G. (1993).
Second order derivatives for network pruning: Optimal brain surgeon.
\emph{Advances in Neural Information Processing Systems (NeurIPS)}.

\bibitem[Louizos et al.(2018)]{louizos2017learning}
Louizos, C., Welling, M., and Kingma, D.~P. (2018).
Learning sparse neural networks through $L_0$ regularization.
\emph{International Conference on Learning Representations (ICLR)}.

\bibitem[Zhu and Gupta(2017)]{zhu2017prune}
Zhu, M. and Gupta, S. (2017).
To prune, or not to prune: Exploring the efficacy of pruning for model compression.
\emph{arXiv preprint arXiv:1710.01878}.

\bibitem[Hanson and Pratt(1988)]{hanson1988comparing}
Hanson, S.~J. and Pratt, L.~Y. (1988).
Comparing biases for minimal network construction with back-propagation.
\emph{Advances in Neural Information Processing Systems (NeurIPS)}, volume~1.

\bibitem[Mozer and Smolensky(1989)]{mozer1989using}
Mozer, M.~C. and Smolensky, P. (1989).
Using relevance to reduce network size automatically.
\emph{Connection Science}, 1(1):3--16.

\bibitem[Li et al.(2019)]{li2019compressing}
Li, T., Wu, B., Yang, Y., Fan, Y., Zhang, Y., and Liu, W. (2019).
Compressing convolutional neural networks via factorized convolutional filters.
\emph{Proceedings of the IEEE/CVF Conference on Computer Vision and Pattern Recognition}.

\bibitem[Li et al.(2017)]{li2017pruning}
Li, H., Kadav, A., Durdanovic, I., Samet, H., and Graf, H.~P. (2017).
Pruning filters for efficient convnets.
\emph{International Conference on Learning Representations (ICLR)}.

\bibitem[He et al.(2018)]{he2018soft}
He, Y., Kang, G., Dong, X., Fu, Y., and Yang, Y. (2018).
Soft filter pruning for accelerating deep convolutional neural networks.
\emph{International Joint Conference on Artificial Intelligence (IJCAI)}, pages 2234--2240.

\bibitem[Mocanu et al.(2018)]{mocanu2018scalable}
Mocanu, D.~C., Mocanu, E., Stone, P., Nguyen, P.~H., Gibescu, M., and Liotta, A. (2018).
Scalable training of artificial neural networks with adaptive sparse connectivity inspired by network science.
\emph{Nature Communications}, 9(1).

\bibitem[Evci et al.(2022)]{evci2022gradient}
Evci, U., Ioannou, Y.~A., Keskin, C., and Dauphin, Y. (2022).
Gradient flow in sparse neural networks and how lottery tickets win.
\emph{AAAI Conference on Artificial Intelligence}.

\bibitem[Zhang et al.(2021)]{zhang2021why}
Zhang, S., Wang, M., Liu, S., Chen, P.-Y., and Xiong, J. (2021).
Why lottery ticket wins? A theoretical perspective of sample complexity on pruned neural networks.
\emph{Advances in Neural Information Processing Systems (NeurIPS)}.

\bibitem[Chen et al.(2021)]{chen2021unified}
Chen, T., Sui, Y., Chen, X., Zhang, A., and Wang, Z. (2021).
A unified lottery ticket hypothesis for graph neural networks.
\emph{International Conference on Machine Learning (ICML)}.

\bibitem[Frantar and Alistarh(2023)]{frantar2023sparsegpt}
Frantar, E. and Alistarh, D. (2023).
SparseGPT: Massive language models can be accurately pruned in one-shot.
\emph{International Conference on Machine Learning}, PMLR.

\bibitem[Sun et al.(2023)]{sun2023simple}
Sun, M., Liu, Z., Bair, A., and Kolter, J.~Z. (2023).
A simple and effective pruning approach for large language models.
\emph{arXiv preprint arXiv:2306.11695}.

\bibitem[Li et al.(2023)]{li2023losparse}
Li, Y., Yu, Y., Zhang, Q., Liang, C., He, P., Chen, W., and Zhao, T. (2023).
LoSparse: Structured compression of large language models based on low-rank and sparse approximation.
\emph{International Conference on Machine Learning (ICML)}, volume 202, pages 20336--20350. PMLR.

\bibitem[Saxe et al.(2019)]{saxe2019mathematical}
Saxe, A.~M., Nelli, S., and Summerfield, C. (2019).
A mathematical theory of semantic development in deep neural networks.
\emph{Proceedings of the National Academy of Sciences}, 116(23):11537--11546.

\bibitem[Hofbauer and Sigmund(1998)]{hofbauer1998evolutionary}
Hofbauer, J. and Sigmund, K. (1998).
\emph{Evolutionary Games and Population Dynamics}.
Cambridge University Press.

\bibitem[Taylor and Jonker(1978)]{taylor1978evolutionary}
Taylor, P.~D. and Jonker, L.~B. (1978).
Evolutionary stable strategies and game dynamics.
\emph{Mathematical Biosciences}, 40(1-2):145--156.

\bibitem[Sandholm(2010)]{sandholm2010population}
Sandholm, W.~H. (2010).
\emph{Population Games and Evolutionary Dynamics}.
MIT Press.

\bibitem[Gunasekar et al.(2018)]{gunasekar2018implicit}
Gunasekar, S., Lee, J.~D., Soudry, D., and Srebro, N. (2018).
Implicit regularization in matrix factorization.
\emph{Advances in Neural Information Processing Systems (NeurIPS)}.

\bibitem[Rumelhart et al.(1986)]{rumelhart1986learning}
Rumelhart, D.~E., Hinton, G.~E., and Williams, R.~J. (1986).
Learning representations by back-propagating errors.
\emph{Nature}, 323(6088):533--536.

\bibitem[Bottou(2010)]{bottou2010large}
Bottou, L. (2010).
Large-scale machine learning with stochastic gradient descent.
\emph{Proceedings of COMPSTAT}.

\bibitem[Li et al.(2018)]{li2018visualizing}
Li, H., Xu, Z., Taylor, G., Studer, C., and Goldstein, T. (2018).
Visualizing the loss landscape of neural nets.
\emph{Advances in Neural Information Processing Systems (NeurIPS)}.

\bibitem[Wen et al.(2016)]{wen2016learning}
Wen, W., Wu, C., Wang, Y., Chen, Y., and Li, H. (2016).
Learning structured sparsity in deep neural networks.
\emph{Advances in Neural Information Processing Systems (NeurIPS)}.

\bibitem[Liu et al.(2017)]{liu2017learning}
Liu, Z., Li, J., Shen, Z., Huang, G., Yan, S., and Zhang, C. (2017).
Learning efficient convolutional networks through network slimming.
\emph{International Conference on Computer Vision (ICCV)}.

\bibitem[LeCun et al.(1998)]{lecun1998gradient}
LeCun, Y., Bottou, L., Bengio, Y., and Haffner, P. (1998).
Gradient-based learning applied to document recognition.
\emph{Proceedings of the IEEE}, 86(11):2278--2324.

\end{thebibliography}

\end{document}